\documentclass[a4paper]{svproc}
\usepackage{url}
\usepackage{graphicx}
\usepackage{subfig}
\usepackage{xcolor}%
\usepackage{array}
\usepackage{amsmath}

\begin{document}
\mainmatter              
\title{Off-Road LiDAR Intensity Based Semantic Segmentation}
\titlerunning{Off-Road LiDAR Intensity Based Semantic Segmentation}  
%
\author{Kasi Viswanath \inst{1}, Peng Jiang \inst{2}, Sujit PB \inst{1}, Srikanth Saripalli \inst{2} }
\authorrunning{Viswanath et al.} 
%
\institute{Department of Electrical Engineering and Computer Science, IISER Bhopal,\\
\email{(kasi18,sujit)@iiserb.ac.in}\\
\and
Department of Mechanical Engineering, Texas A$\&$M University,\\
\email{(maskjp,ssaripalli)@tamu.edu}}
\maketitle              

\begin{abstract}
LiDAR is used in autonomous driving to provide 3D spatial information and enable accurate perception in off-road environments, aiding in obstacle detection, mapping, and path planning. Learning-based LiDAR semantic segmentation utilizes machine learning techniques to automatically classify objects and regions in LiDAR point clouds. Learning-based models struggle in off-road environments due to the presence of diverse objects with varying colors, textures, and undefined boundaries, which can lead to difficulties in accurately classifying and segmenting objects using traditional geometric-based features. In this paper, we address this problem by harnessing the LiDAR intensity parameter to enhance object segmentation in off-road environments. Our approach was evaluated in the RELLIS-3D data set and yielded promising results as a preliminary analysis with improved mIoU for classes "puddle" and "grass" compared to more complex deep learning-based benchmarks\footnote{https://github.com/MOONLABIISERB/lidar-intensity-predictor/tree/main}. The methodology was evaluated for compatibility across both Velodyne and Ouster LiDAR systems, assuring its cross-platform applicability. This analysis advocates for the incorporation of calibrated intensity as a supplementary input, aiming to enhance the prediction accuracy of learning based semantic segmentation frameworks.
\footnote{The work is supported by \textbf{TIH iHUB Drishti-IIT Jodhpur} under project number \textbf{23} and accepted for publication at International Symposium on Experimental Robotics 2023. }
\end{abstract}
\section{Introduction}
Off-road autonomous driving has recently received much attention with multimodal datasets, improved semantic segmentation frameworks and robust planners. Multiple sensors are employed to perceive and navigate unstructured terrain. The Light Detection and Ranging (LiDAR) sensor provides 3D information that helps to extract geometric data about a scene, while a camera captures visual data. However, cameras can lose information at different lighting conditions, which can affect navigation algorithms. To solve this problem, we used the backscattered LiDAR intensity value, which is not affected by lighting conditions, to classify objects in an off-road environment.

Traditional LiDAR Semantic Segmentation models \cite{10.1007/978-3-030-64559-5_16} \cite{9010002} have focused primarily on leveraging the geometric properties of objects, which has proven effective for urban scenes characterized by well-defined boundaries. However, off-road scenes present unique challenges, as they contain diverse objects/classes with varying colors, textures, and undefined boundaries. In such scenarios, the aforementioned models may not perform optimally.

The use of LiDAR intensity as an auxiliary input in conjunction with geometry information has been previously explored for segmentation purposes \cite{salsanet2020}. However, the values of the LiDAR intensity are influenced by factors such as range, angle of incidence, and surface reflectivity. In this paper, we explore the use of surface reflectivity information for efficiently segmenting different classes for off-road scenes by calibrating intensity values for range and angle of incidence.

\begin{figure}[!ht]
    \centering
    \includegraphics[width = \textwidth,height = 4 cm]{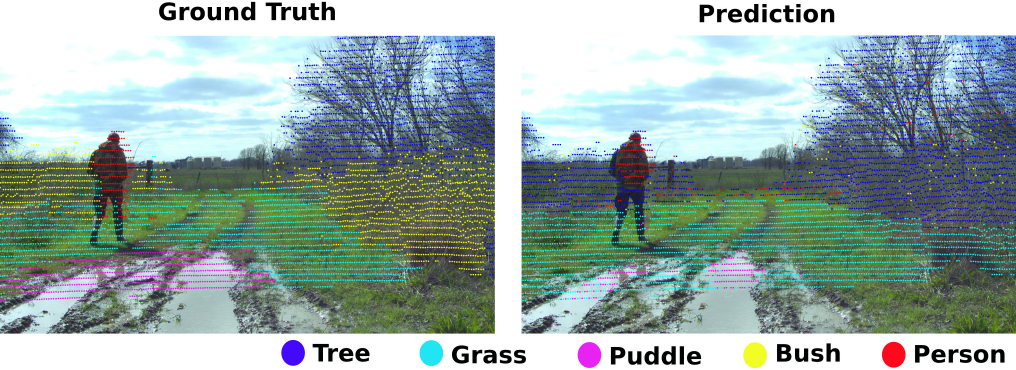}
    \caption{Segmentation Results on RELLIS-3D dataset.}
    \label{fig:puddle}
\end{figure}


\section{LiDAR Intensity Calibration}
The intensity value in a LiDAR denotes the returned backscattered signal ebergy. 
Intensity is dependent on multiple factors such as range ($R$), reflectivity of the object ($\rho$), angle of incidence ($\alpha$), surface roughness, atmospheric humidity, etc.\cite{weitkamp_lidar_2006}. 
Among these factors, the surface's reflectivity is predominantly tied to the object's properties, making it our focal parameter. To ensure a reliable LiDAR intensity-based classification, it is crucial to calibrate the intensity values for the dependencies on the range and angle of incidence.

\subsection{Range Dependence}
LiDAR sensor's principle works on the difference between the emitted and received laser power, given by: 
\begin{equation}
    \centering
    P_r = \frac{P_e\rho Cos(\alpha)}{R^2}
    \label{lq}
\end{equation}
where $P_e$ is the emitted laser power,  $\rho$ is reflectance of an object. Eq. (\ref{lq}) represents an ideal model in which the intensity is inversely proportional to the square of the range.
\begin{figure}[t]
\vspace{-0.5 cm}
\centering
    \subfloat[]{\label{r_int}\includegraphics[width = 0.45\textwidth,height = 3.4cm]{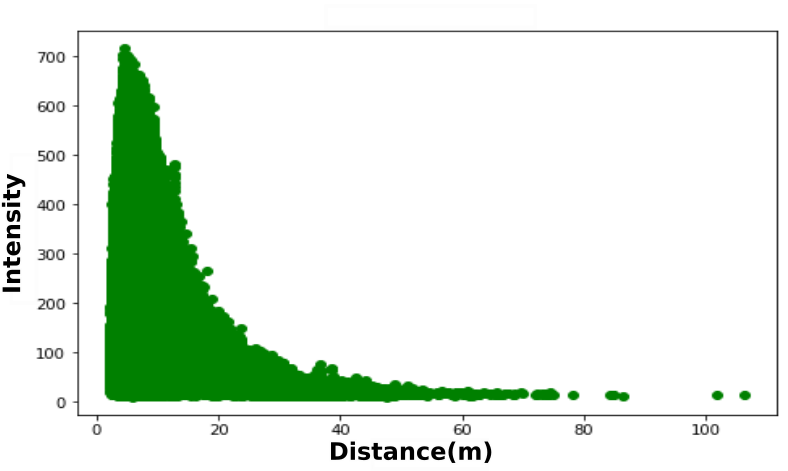}}
    \subfloat[]{\label{c_int}\includegraphics[width = 0.45\textwidth,height = 3.6 cm]{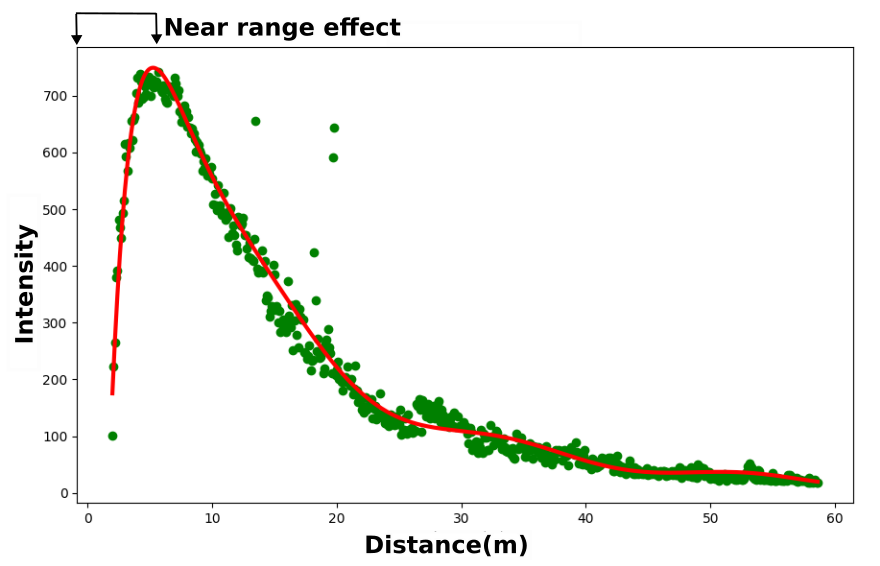}}    
\caption{(a) Raw Intensity vs Range of Grass (b) Intensity calibrated for $\alpha$ vs Range.}
\end{figure}
However, in reality, the range-intensity relationship at proximate distances is non ideal. In a shorter range, the Range-Intensity exhibits an exponential relationship, Eq. (\ref{ir}) until reaching a certain threshold \cite{Biavati:11} as given by:
\begin{equation}
    \centering
    I(R,\alpha,\rho) \propto P_r(R,\alpha,\rho) = \eta(R)\frac{I_e\rho Cos(\alpha)}{R^2}
    \label{ir}
\end{equation}
Fig. \ref{r_int} shows the range vs. intensity scatter plot for grass, including the incidence angle. Due to the near-range effect\cite{rs14174393}, our analysis focuses on points within the range of 6 to 60 m.

\subsubsection{Intensity Correction for Large Range}
To correct the LiDAR intensities that depend solely on an object's reflectance, we follow Eq.\ref{pred_eq} and focus on LiDAR points outside of the near-range effect ($R>6 \text{ meter}$, where $\eta(R) = 1$). 
\begin{equation}
    I(\rho) = \frac{I(R,\alpha,\rho)  R^2}{Cos(\alpha)}
    \label{pred_eq}
\end{equation}
The above computation provides the emitted intensity (originating from the object) adjusted for both range and angle of incidence. Correction for the range is made through Eq (\ref{eq:range_cor}): 
\begin{equation}
    \centering
    I_e = I(R,\rho)  R^2\qquad\{R > 6 m\}
    \label{eq:range_cor}
\end{equation}
\subsection{Angle of Incidence Dependence}
\begin{figure}[!ht]
 \centering
    \subfloat[]{\label{fig:alphavsI}\includegraphics[width = 0.45\textwidth, height = 4.3 cm]{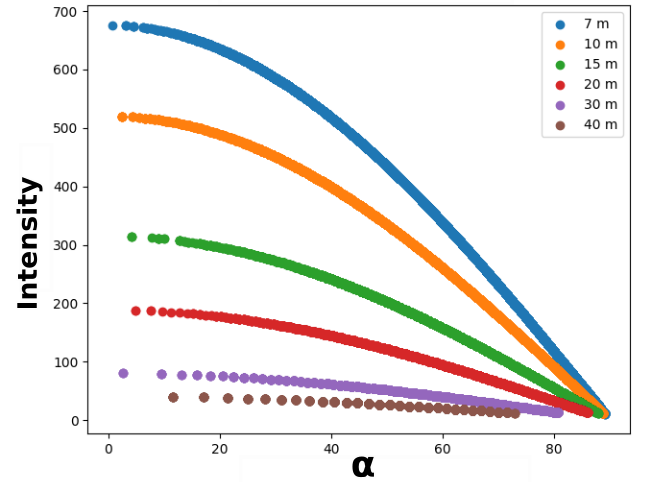}}
    \subfloat[]{\label{fig:avic}\includegraphics[width = 0.45\textwidth,height = 4.3 cm]{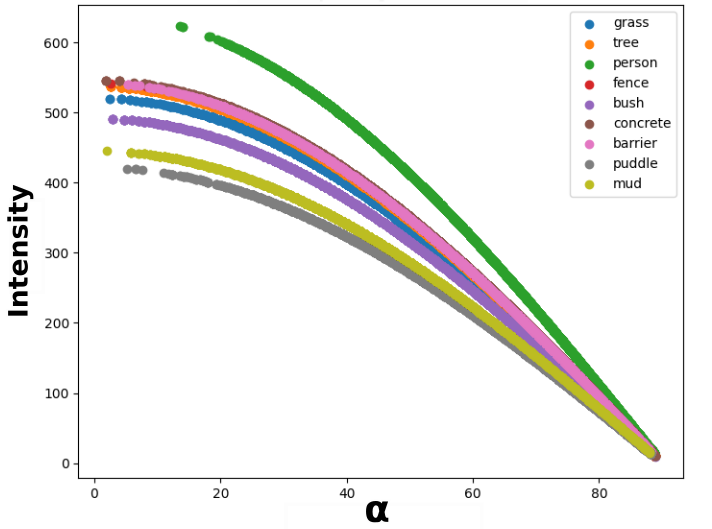}}   
\caption{(a) Raw Intensity vs $\alpha$ of Grass at different ranges (b) Raw intensity vs $\alpha$ for different classes at 10 meters range. The segregation of intensity values for different classes is observed.}
\end{figure}

The intensity is intrinsically linked to angle of incidence ($\alpha$). Fig.\ref{fig:alphavsI} and Fig. \ref{fig:avic} elucidate the relationship between the angle of incidence ($\alpha$) and the Intensity ($I$). The graphs show that the highest intensity is observed when $\alpha$ is close to 0, and the lowest when it is close to $\pi/2$, which is in agreement with the relationship expressed in Eq. (\ref{ir}). 

Based on Eq. (\ref{pred_eq}), the effect of the incidence angle $\alpha$  can be eliminated by dividing the intensity $I(R,\alpha,\rho)$ by $\cos(\alpha)$, resulting in $I(R,\rho)$, which depends only on the range and reflectance of the surface. The corrected intensity versus range plot is illustrated in Fig. \ref{c_int}.
\begin{figure}[t]
    \centering
    \includegraphics[width = 0.8\textwidth,height = 3 cm]{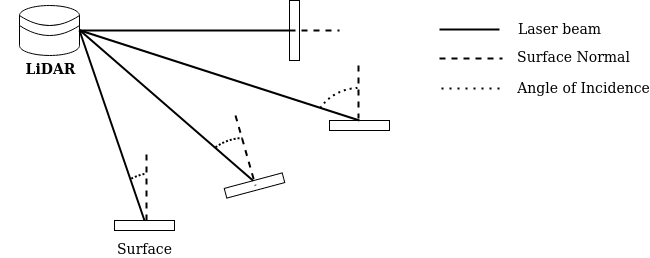}
    \caption{Interaction of laser beam with surface at different angle of incidence }
    \label{fig:alpha_surf}
\end{figure}

The angle $\alpha$ is determined by the vector of the incident laser beam$(\overrightarrow{l})$ and the normal of the surface $(\overrightarrow{n})$ as shown in Fig. \ref{fig:alpha_surf}. It can be calculated as $\alpha = arccos(\overrightarrow{l} \cdot \overrightarrow{n})$.
Traditionally, the surface normal can be determined by Ball query sampling \cite{6287634}, which takes into account points in the vicinity within a certain radius to calculate the normal vector by fitting a plane with Principal Component Analysis (PCA) \cite{doi:10.1080/14786440109462720}. However, the angle of incidence calculated with the estimated normal has an MSE error of 0.44 (38 degrees), which is inaccurate. Therefore, to accurately estimate the angle of incidence, we fit the surface normal-point vector data to the ground truth angle of incidence using Fully Connected Layers(FCN).

For FCN, we need the ground truth that is determined in the following way. The ground truth $\alpha$ is obtained by segregating the point cloud data according to annotated classes and then associating them with their corresponding range information, as shown for the "grass" class in Figure \ref{r_int}. Since $\alpha$ has a $cosine$ relationship with intensity, the maximum intensity in every range in Figure \ref{r_int} will have ($\cos \alpha$) = 1. This allows us to calculate $\alpha$ for a given LiDAR point class using the following equation: 
\begin{equation}
\alpha = arccos(\frac{Intensity(R)}{MaxIntensity(R)} \label{gt_alpha}). 
\end{equation}
The generated ground truth $\alpha$ along with their corresponding surface normal point vector is used to train the FCN. The FCN takes the surface normal point vector (6 element array) as input and predicts $\alpha'$. The mean absolute error (MAE) between $\alpha$ and $\alpha'$ is calculated and the loss is backpropagated for the FCN to learn.

\section{Experiments and Results}
The proposed approach to using surface reflectivity for segmentation is evaluated in the Rellis-3D off-road data set\cite{9561251}. Rellis-3D is a multimodal dataset consisting of 4 sequences of annotated LiDAR point cloud from off-road environments. The dataset consists of 20 classes that include grasses, bushes, puddles, trees, etc., providing heterogeneity. The point cloud data are collected using Ouster OS1 with 64 channels. For initial experimentation, we only consider the major classes such as grass, bushes, trees, puddle, and person for semantic labeling.
\begin{figure}[t]
    \centering
    \includegraphics[width = 0.8\textwidth,height = 4 cm]{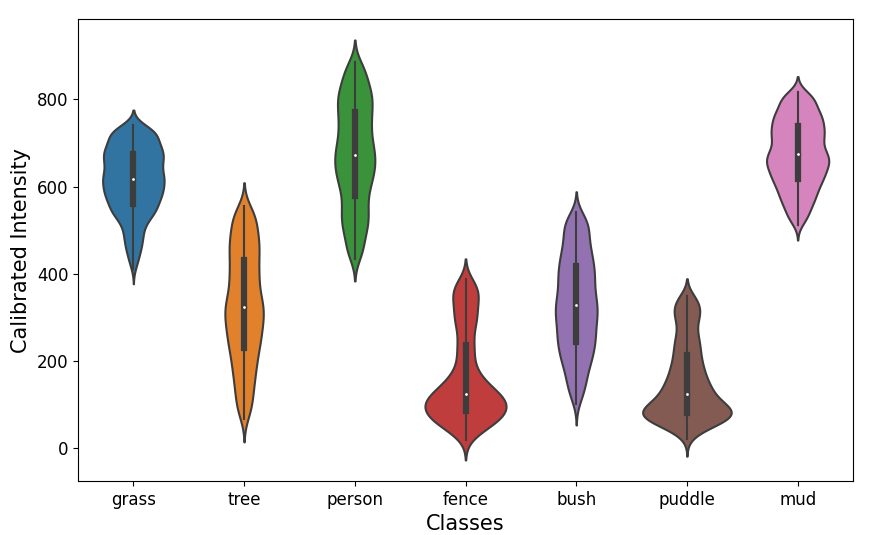}
    \caption{Calibrated Intensity ranges of different classes.}
    \label{fig:int_ranges}
\end{figure}
Ouster OS1 intensity data is purely raw, i.e; the values are not pre-calibrated but are scaled to 64-bit integers. The calibrated intensity ranges for different classes were generated from the 0000 sequence of Ouster data. $\alpha$ for each LiDAR point is extracted using Eq.(\ref{gt_alpha}). With the $\alpha$ values and range known, we calibrate the raw LiDAR intensity using Eq.(\ref{pred_eq}). The calibrated intensity values are segregated based on the annotated labels and we find that the intensity values are distributed in specific ranges for different classes, as shown in Figure \ref{fig:int_ranges}. This proves our hypothesis of reflectivity-based class segregation.

To predict a LiDAR point cloud, the intensity values are calibrated using the Eq. (\ref{pred_eq}) where $\alpha$ is predicted using the FCN. A neighborhood prediction policy is employed, whereby calibrated intensity values are assigned classes based on their proximity to the closest class mode values. The predictions were performed on sequences 0001 and 0002 of the RELLIS-3D dataset, with classes limited to grass, bush, trees, person, and puddle. The framework gave an average mIoU of $47\%$, and the respective class IoU is given in Table \ref{tab:results}. 

\begin{table}
    \centering
    \small
    \begin{tabular}{|c|c|c|c|c|c|c|}
         \hline
         Framework & Tree & Grass & Puddle & Bushes & Person & mean \\
         \hline
         Ours & 74.68 & \textbf{66.44} & \textbf{47.83} & 13.65 & 33.52 & 47.17 \\
         \hline
         SalsaNext$^*$ & \textbf{79.04} & 64.74 & 23.20 & \textbf{72.90} & \textbf{83.17} & \textbf{64.61}\\
         \hline
         KPConv$^*$ & 49.25 & 56.41 & 0.0 & 58.45 & 81.20 & 49.06  \\
         \hline
    \end{tabular}
    \vspace{0.2 cm}
    \caption{mIoU of experiment results. SalsaNext\cite{10.1007/978-3-030-64559-5_16} and KPConv\cite{9010002} are benchmarks of RELLIS-3D dataset.}
    \label{tab:results}
\end{table}
\vspace{-1 cm}
\subsection{Pre-Processing Velodyne Intensity data.}
\begin{figure}[!ht]
    \centering
    \includegraphics[width = 0.8\textwidth]{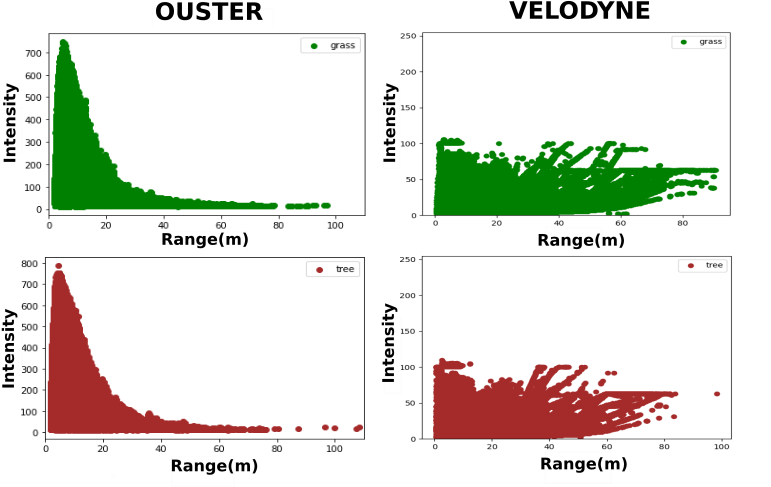}
    \caption{Raw intensity plots from Ouster and Velodyne LiDAR for grass(green) and tree(brown).}
    \label{fig:raw_osvs}
\end{figure}

\begin{figure}
    \centering
    \includegraphics[width = \textwidth, height = 4 cm]{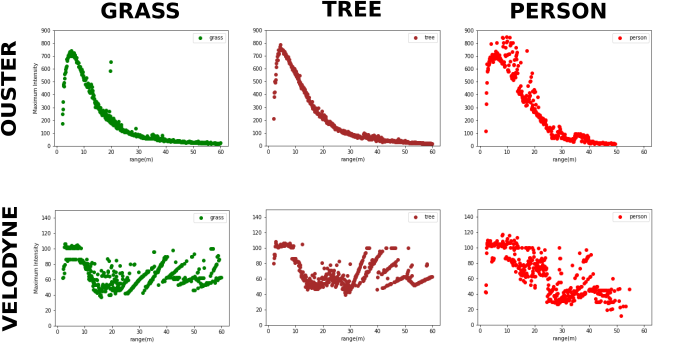}
    \caption{Maximum intensity vs range for classes Grass, Tree and Person from Ouster and Velodyne LiDAR scans.}
    \label{fig:max_osvs}
\end{figure}

The Velodyne LiDARs generate point cloud data comprising three-dimensional coordinate points and a preprocessed intensity dataset, which differs from the raw intensity data produced by Ouster LiDARs. The Velodyne Intensity data are adjusted to an 8-bit integer scale and are calibrated with respect to range and laser power, as detailed in the data sheet. Figure \ref{fig:raw_osvs} illustrates a comparison between the intensity data from Velodyne and Ouster LiDARs across various classes. As depicted in the figure, the Velodyne intensity data do not align with the LiDAR equation \ref{ir}. In this context, we introduce a method for converting the Velodyne intensity data back to the raw intensity format used by Ouster, essentially reversing the inherent intensity processing employed by Velodyne.
The RELLIS-3D dataset contains LiDAR scans that are annotated, and these scans are obtained using the 32-channel Velodyne Ultra Puck for the same scenes as those scanned by Ouster LiDARs. In Figure \ref{fig:max_osvs}, we illustrate the extraction of the most intense values that correspond to specific ranges for each class, and this is done using 1000 scans from both Velodyne and Ouster. It is worth noting that Velodyne LiDARs, as per the datasheet, lack calibration for the angle of incidence, denoted by [$\alpha$]. This lack of calibration means that the maximum intensity value remains independent of the angle of incidence, which is an important point to consider.

\begin{figure}[!h]
    \centering
    \includegraphics[width = \textwidth, height = 6 cm]{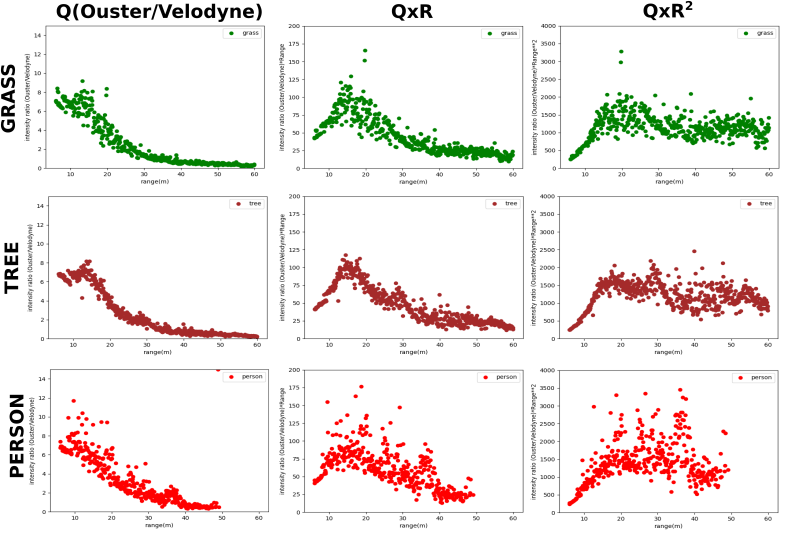}
    \caption{$Q$, $Q*R, Q*R^2$ function for class Grass, Tree and Person.}
    \label{fig:q}
\end{figure}

In Figure \ref{fig:q}, we can see the division of the maximum intensity values of Ouster by the maximum intensity values of Velodyne, resulting in a ratio denoted [$Q$]. It is noticeable that the trends or slopes for the three classes(grass, tree, and person) are strikingly similar. To explore further characteristics, we multiply [$Q$] by the range [$R$] and [$R^2$]. 

\begin{figure}
    \centering
    \includegraphics[width = \textwidth, height = 4.5 cm]{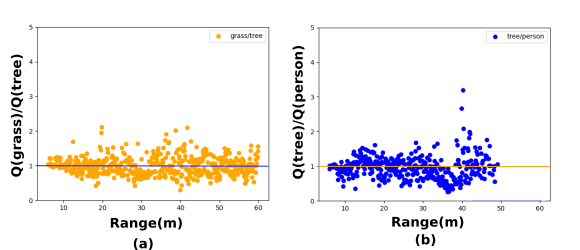}
    \caption{Comparing [Q] values between different classes to confirm object-independence of calibration function}
    \label{fig:q_r}
\end{figure}

To further validate the independence of the calibration function from objects, we performed a comparison by dividing the [$Q$] values of different classes with each other. Figure \ref{fig:q_r} presents the deviations in the [$Q$] values, resulting in an average value of 1. This outcome provides strong confirmation that [$Q$] remains unchanged by variations in the reflectivity parameter, underscoring its object-independent nature.

We apply a polynomial fitting procedure to the function $Q(r)$. This allows us to transform Velodyne intensity data into the Ouster format by utilizing the following equation:
\begin{equation}
\centering
\text{Raw Intensity} = Q(r) \times \text{Velodyne}(r)
\end{equation}
This method facilitates a seamless conversion of classification ranges from Ouster to Velodyne, enabling a direct translation between the two.
\begin{figure}
    \centering
    \includegraphics[width = \textwidth]{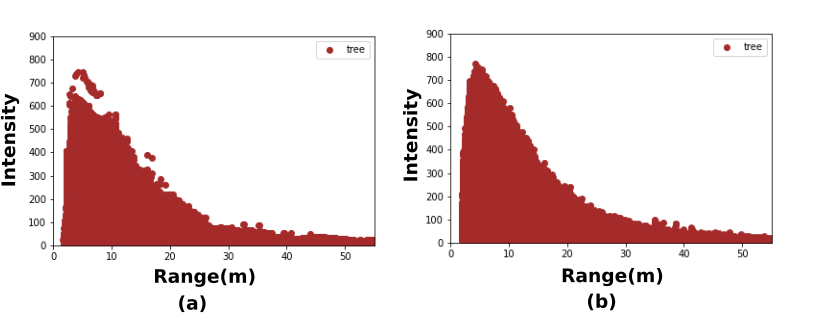}
    \caption{(a)Velodyne Intensity data of class tree after pre-processing (b) Ouster intensity data of class Tree.}
    \label{fig:enter-label}
\end{figure}
\section{Experimental Insights}

The decreased mIoU score for the "bush" class can be attributed to the considerable overlap in the calibrated intensity range with the "tree" class. This is likely due to the similar texture of leaves, stems, and trunks shared by both bushes and trees within the dataset.

During testing, we have observed that the class "puddle" is predicted with significantly more accurate distinct boundaries compared to the ground truth, as shown in Fig.\ref{fig:puddle}. This improvement is achieved by using a mode-based prediction approach that effectively removes the major outliers from the segmentation distribution.

LiDAR scans with higher point density (more channels) yield more accurate surface normal estimations than scans with lower point density. This finding was also observed in \cite{10.1117/12.2663098}. We noticed that the approach yields better predictions for ranges beyond 10 meters. This improvement can be attributed to the more accurate estimation of $\alpha$ at larger distances, leading to precise estimates of calibrated intensity. 

\section{Conclusions}
 In this paper we presented analysis of the potential utilization of LiDAR intensity for semantic segmentation of terrestrial LiDAR scans.  Our proposed pipeline has demonstrated superior performance compared to the RELLIS-3D benchmarks, particularly in the prediction of the "puddle" and "grass" classes. It is worth noting that, while our approach may exhibit a lower prediction accuracy compared to other learning-based segmentation frameworks, it is important to emphasize that this study serves as an initial exploration into harnessing the LiDAR intensity parameter for this specific task. We also introduce a preprocessing methodology tailored for Velodyne LiDARs, aiming to enhance cross-platform compatibility. We acknowledge that the current methodology's accuracy can potentially be enhanced by incorporating geometric information derived from LiDAR points or by integrating sparse semantic data from camera images, thereby addressing the challenge of calibrated range overlap. Further investigations can be conducted to test the efficacy of the methods for different climatic conditions as the vegetation and texture of an off-road scene change significantly compared to urban environments.
%
%
\bibliographystyle{splncs03_unsrt}
\bibliography{myReference}

\begin{thebibliography}{10}
\providecommand{\url}[1]{\texttt{#1}}
\providecommand{\urlprefix}{URL }

\bibitem{10.1007/978-3-030-64559-5_16}
Cortinhal, T., Tzelepis, G., Erdal~Aksoy, E.: Salsanext: Fast,
  uncertainty-aware semantic segmentation of lidar point clouds. In: Bebis, G.,
  Yin, Z., Kim, E., Bender, J., Subr, K., Kwon, B.C., Zhao, J., Kalkofen, D.,
  Baciu, G. (eds.) Advances in Visual Computing. pp. 207--222. Springer
  International Publishing, Cham (2020)

\bibitem{9010002}
Thomas, H., Qi, C.R., Deschaud, J.E., Marcotegui, B., Goulette, F., Guibas, L.:
  Kpconv: Flexible and deformable convolution for point clouds. In: IEEE/CVF
  International Conference on Computer Vision. pp. 6410--6419 (2019)

\bibitem{salsanet2020}
Aksoy, E.E., Baci, S., Cavdar, S.: Salsanet: Fast road and vehicle segmentation
  in lidar point clouds for autonomous driving. In: IEEE Intelligent Vehicles
  Symposium (IV2020) (2020)

\bibitem{weitkamp_lidar_2006}
Weitkamp, C.: Lidar: range-resolved optical remote sensing of the atmosphere,
  vol. 102. Springer Science \& Business

\bibitem{Biavati:11}
Biavati, G., Donfrancesco, G.D., Cairo, F., Feist, D.G.: Correction scheme for
  close-range lidar returns. Appl. Opt.  50(30),  5872--5882 (Oct 2011),
  \url{https://opg.optica.org/ao/abstract.cfm?URI=ao-50-30-5872}

\bibitem{rs14174393}
Cheng, Y.T., Lin, Y.C., Habib, A.: Generalized lidar intensity normalization
  and its positive impact on geometric and learning-based lane marking
  detection. Remote Sensing  14(17) (2022),
  \url{https://www.mdpi.com/2072-4292/14/17/4393}

\bibitem{6287634}
Shakhnarovich, G., Darrell, T., Indyk, P.: New Algorithms for Efficient
  High-Dimensional Nonparametric Classification, pp. 75--101 (2006)

\bibitem{doi:10.1080/14786440109462720}
F.R.S., K.P.: Liii. on lines and planes of closest fit to systems of points in
  space. The London, Edinburgh, and Dublin Philosophical Magazine and Journal
  of Science  2(11),  559--572 (1901)

\bibitem{9561251}
Jiang, P., Osteen, P., Wigness, M., Saripalli, S.: Rellis-3d dataset: Data,
  benchmarks and analysis. In: IEEE International Conference on Robotics and
  Automation (ICRA). pp. 1110--1116 (2021)

\bibitem{10.1117/12.2663098}
Yu, J., Chen, J., Dabbiru, L., Goodin, C.T.: {Analysis of LiDAR configurations
  on off-road semantic segmentation performance}. In: Autonomous Systems:
  Sensors, Processing, and Security for Ground, Air, Sea, and Space Vehicles
  and Infrastructure. vol. 12540, p. 1254003. SPIE (2023),
  \url{https://doi.org/10.1117/12.2663098}

\end{thebibliography}

\end{document}